\documentclass{article}
\usepackage{pdfpages}
\usepackage[english]{babel}
\usepackage{float}
\usepackage[letterpaper,top=2cm,bottom=2cm,left=3cm,right=3cm,marginparwidth=1.75cm]{geometry}

\usepackage{pifont}
\usepackage{rotating}
\usepackage{makecell}
\usepackage{array}
\usepackage{multirow}
\usepackage{multicol}
\usepackage{booktabs}
\usepackage{amsmath}
\usepackage{graphicx}
\usepackage{bbding}
\usepackage[colorlinks=true, allcolors=blue]{hyperref}
\usepackage{pdflscape}
\usepackage{url}

\usepackage{fancyhdr}

\fancypagestyle{plain}{
    \fancyhf{}
    \fancyhead[L]{\includegraphics[height=0.8cm]{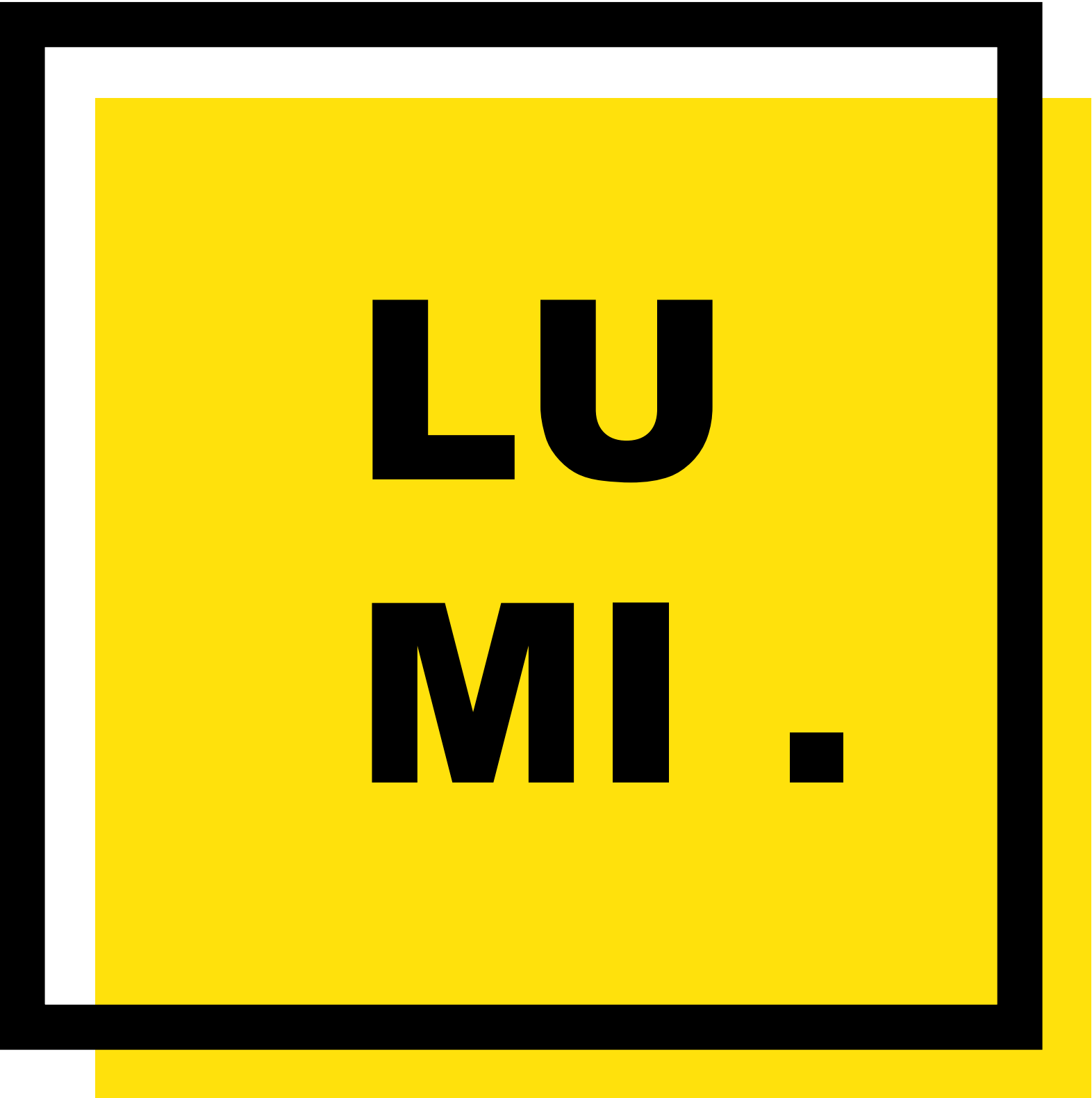}}
    
}

\title{Towards Real-World Ultrasound Understanding: Large Vision-Language Models from Multi-Image Examinations with Long-Form Reports}

\author{
Bingcong Yan$^{1}$,
Chunlei Li$^{1}$,
Jingliang Hu$^{1}$,
Yilei Shi$^{1}$,
Xiao Xiang Zhu$^{2}$,
Lichao Mou$^{1,*}$
}

\date{}

\begin{document}
\maketitle

\begin{center}
\small
$^{1}$MedAI Technology (Wuxi) Co. Ltd.\\
$^{2}$Technical University of Munich\\[0.5em]
$^{*}$Correspondence: lichao.mou@gmail.com
\end{center}

\begin{center}
\url{https://medai-t.github.io/LUMI/}
\end{center}

\begin{abstract}
Large vision-language models (LVLMs) have achieved strong performance across many medical imaging tasks, yet their application to ultrasound remains limited due to its inherent complexity and variability. In this work, we revisit what is truly needed to enable real-world ultrasound understanding. Instead of introducing complex architectures or elaborate training strategies, we show that data scale and clinically faithful data alignment are the key factors. We construct a large-scale dataset of 1.5M real-world ultrasound examinations, containing 17.7M images, multi-organ coverage, and paired uncurated clinical reports. Crucially, we organize the data at the examination level, aligning multiple images with their corresponding reports to reflect real clinical workflows. We then fine-tune a standard LVLM using low-rank adaptation (LoRA) on this dataset without task-specific modifications. Surprisingly, this simple recipe already leads to strong performance across diverse ultrasound understanding tasks, outperforming prior methods designed with more complex pipelines. Beyond these results, we present model and data scaling analyses that provide insights into the role of scale in ultrasound LVLMs.
\end{abstract}

\section{Introduction}
Large vision-language models (LVLMs) have recently demonstrated remarkable performance across both general-domain \cite{DBLP:journals/corr/abs-2508-18265,DBLP:journals/corr/abs-2511-21631,DBLP:journals/corr/abs-2504-07491,DBLP:journals/tmlr/0080ZGZ00ZZL0L25} and medical imaging tasks \cite{DBLP:journals/corr/abs-2406-19280,DBLP:journals/corr/abs-2506-07044,DBLP:journals/corr/abs-2604-05081,DBLP:journals/corr/abs-2309-09812,DBLP:conf/nips/LiWZULYNPG23,DBLP:conf/ml4h/MoorHWYDLZRR23,DBLP:journals/corr/abs-2506-15477,lin2025healthgpt}, including computed tomography (CT), magnetic resonance imaging (MRI), X-ray, and positron emission tomography (PET). By jointly modeling visual content and textual descriptions, these models enable a wide range of applications such as visual question answering, report generation, and clinical decision support.

Despite this progress, extending LVLMs to ultrasound remains highly challenging. Unlike CT, MRI, and X-ray, which provide relatively standardized and global anatomical views, ultrasound has a limited field of view and is inherently examination-centric. In routine clinical practice, a single ultrasound examination typically consists of multiple images acquired from different viewpoints, often covering multiple organs. Even for a single organ, several views are usually required for comprehensive assessment. Furthermore, ultrasound imaging is highly operator-dependent and frequently affected by noise, artifacts, acoustic shadowing, and varying acquisition protocols, leading to substantially greater complexity and variability compared with other imaging modalities.

Several recent studies have begun exploring LVLMs for ultrasound understanding\footnote{We note that recent work has also explored other types of foundation models for ultrasound imaging, including vision foundation models \cite{DBLP:journals/mia/JiaoZLXHHWZZWG24,DBLP:journals/corr/abs-2509-11752} and CLIP-style models \cite{
Christensen2024VisionlanguageFM,Vukadinovic2025ComprehensiveEE,DBLP:journals/corr/abs-2502-14807,Guo2026AVG,DBLP:journals/corr/abs-2604-01749}. While complementary to LVLMs, these approaches fall outside the scope of this work and are not discussed in detail here.}. For example, \cite{DBLP:journals/corr/abs-2509-25748} fine-tunes Qwen2.5-VL using ultrasound textbooks and image-text pairs, primarily based on single-image captioning data. Similarly, \cite{DBLP:conf/mir/GuoSWCLT25} trains a multimodal large language model (MLLM) for ultrasound image captioning and question answering. \cite{DBLP:journals/corr/abs-2506-07837} investigates chain-of-thought (CoT) fine-tuning of Qwen2.5-VL-7B-Instruct for ultrasound reasoning at the image level. \cite{DBLP:journals/corr/abs-2511-22256} extends a base LVLM to support not only captioning but also segmentation and detection, albeit still within single-image settings. \cite{DBLP:journals/corr/abs-2509-14977} further incorporates organ-level image-text pairs with multiple images per sample to train a customized Qwen2-VL model. While these efforts demonstrate promising early results, they are still developed under relatively simplified settings, focusing on either single images or single organs. In real-world clinical workflows, ultrasound examinations naturally involve complex relationships among multiple images, multiple organs, and long-form reports. For instance, abdominal ultrasound examinations routinely include the liver, gallbladder, pancreas, spleen, and kidneys, with different organs often co-occurring within the same image. 

The discrepancy between existing training settings and practical clinical scenarios remains largely underexplored. More fundamentally, an open question is whether advancing ultrasound LVLMs truly requires sophisticated model architectures or whether data scale and clinically grounded data alignment play a more decisive role.

In this work, we take a step toward real-world ultrasound understanding by training an LVLM on large-scale ultrasound data collected in the wild. Our framework leverages clinically faithful examination-level supervision, where multiple images from the same examination are paired with their corresponding clinical report. By modeling the relationships between these images and the associated report, the model learns to perform more realistic and comprehensive ultrasound reasoning.

Our contributions are summarized as follows:
\begin{itemize}
    \item We move beyond image-level or organ-level supervision and study LVLMs under realistic clinical scenarios involving multi-image examinations and rich long-form textual supervision.
    \item We construct a large-scale ultrasound dataset containing nearly 1.5 million examinations, 17.7 million images, over 14 organs, and extensive clinical reports collected from real-world practice.
    \item We train an LVLM on this dataset, which we call LUMI, and benchmark a broad range of large models, achieving state-of-the-art performance.
\end{itemize}

\section{Dataset}
\label{sec:dataset}
\subsection{Data Collection}
\label{subsec:data-collection}
\subsubsection{Sources and Composition}
\label{subsubsec:sources-composition}
Our dataset consists of de-identified ultrasound examinations collected from multiple large medical institutions between 2014 and 2025. The data is organized at the examination level, where each case contains the complete set of ultrasound images together with the corresponding detailed clinical report.

The dataset covers five major clinical domains, including thyroid, breast, upper abdomen, gynecology, and male urinary examinations. The examinations involve a diverse set of anatomical structures, including the liver, gallbladder, spleen, pancreas, thyroid, cervical lymph nodes, axillary lymph nodes, breast, kidneys, uterus, ovaries, bladder, prostate, and carotid arteries, among others. A key characteristic of real-world ultrasound imaging is the frequent co-occurrence of multiple organs within the same scanning plane, which is naturally preserved in our dataset. For example, the liver and gallbladder commonly appear together in upper abdominal scans, while the uterus and ovaries are often jointly visualized in gynecological examinations.

\subsubsection{Curation and Preprocessing}
\label{subsubsec:curation-preprocessing}
To ensure data quality, consistency, and privacy compliance, we applied a multi-stage curation and preprocessing pipeline:
\begin{itemize}
    \item \textbf{Privacy and anonymization.} All personally identifiable information (PII) was rigorously removed from both imaging data and textual reports, including patient names, identification numbers, and contact information.
    \item \textbf{Case filtering.} We excluded cases with missing images or reports. Cases containing an anomalously large number of images ($>32$) were removed. We further discarded reports that were excessively short (length $<5$ characters), which typically provide insufficient clinical information, as well as excessively long reports (length $>3000$ characters), which often contain redundant or non-standard content.
    \item \textbf{Standardization.} All ultrasound images were converted into a unified format and resolution to ensure consistency across institutions and acquisition devices. Textual reports were further cleaned by removing institutional metadata, such as hospital and department identifiers, followed by formatting normalization and tokenization.
\end{itemize}

\subsection{Dataset Statistics}
\label{subsec:dataset-statistics}
\subsubsection{Scale and Basic Demographics}
\label{subsubsec:scale-demographics}
\begin{figure}[t]
\centering
\includegraphics[width=1.0\linewidth]{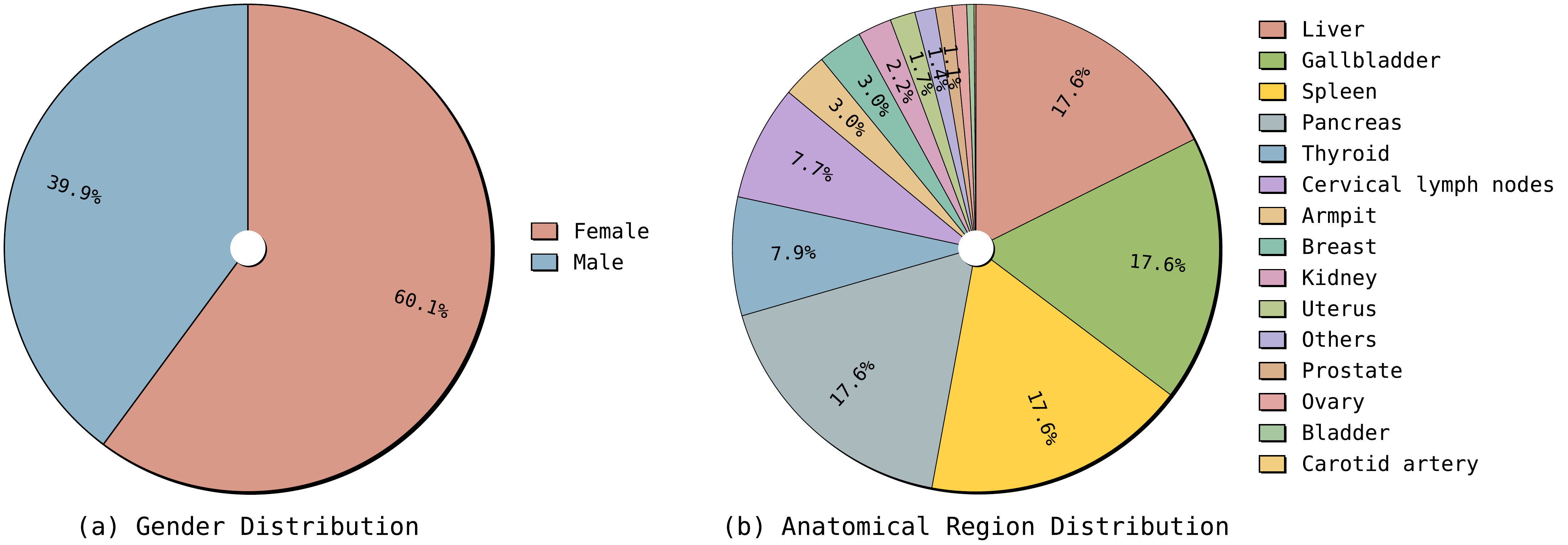}
\caption{Demographic and anatomical distributions of the ultrasound dataset. (a) Patient gender distribution. (b) Distribution of anatomical structures covered by the examinations.}
\label{fig:gender_anatomical_regions-distribution}
\end{figure}

\begin{figure}[t]
\centering
\includegraphics[width=0.8\linewidth]{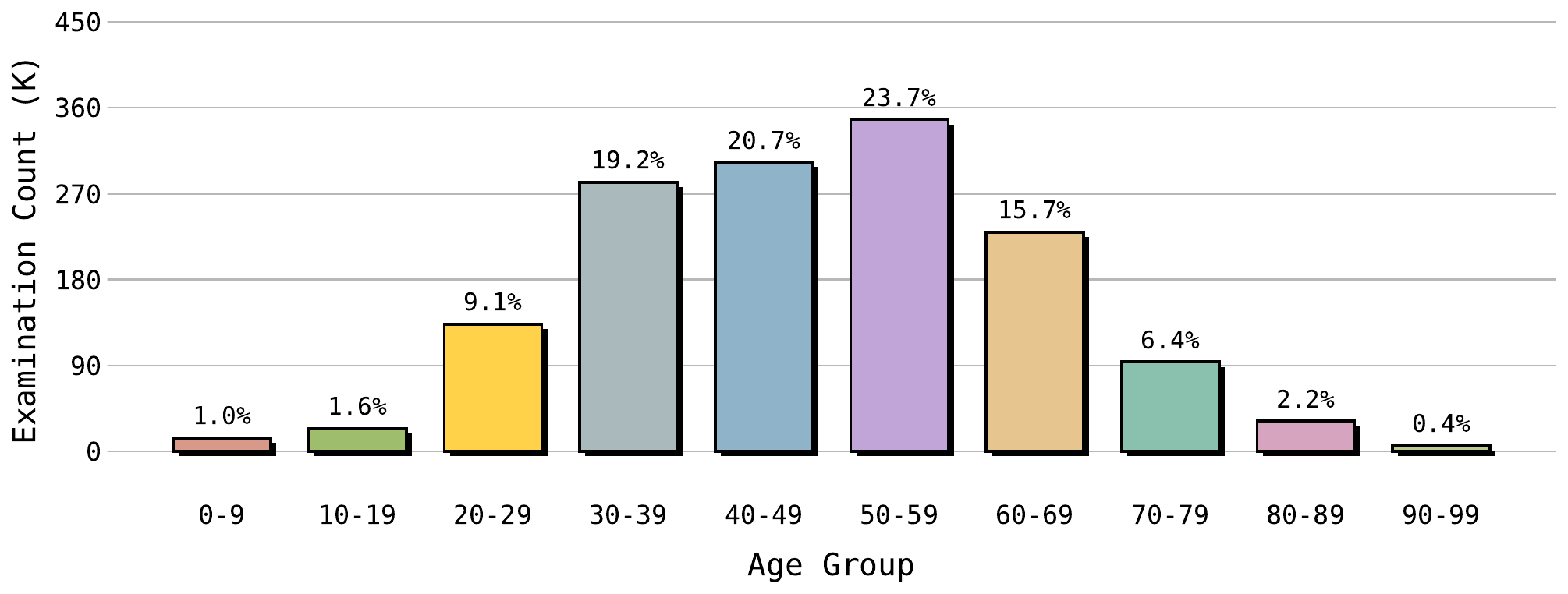}
\caption{Age distribution of patients in the dataset.}
\label{fig:age-distribution}
\end{figure}

The final curated dataset contains 1,464,461 examination cases and 17,673,019 ultrasound images, with an average of 12.1 images per examination. The gender distribution of the patient cohort is illustrated in Figure~\ref{fig:gender_anatomical_regions-distribution}, while the age distribution is shown in Figure~\ref{fig:age-distribution}.

\subsubsection{Examination Distribution}
\label{subsubsec:exam-distribution}
The distribution of examination categories is presented in Figure~\ref{fig:gender_anatomical_regions-distribution}. Upper abdomen examinations constitute the largest portion of the dataset, followed by thyroid examinations. Breast, gynecology, and male urinary examinations are also well represented, providing broad clinical coverage for model training and evaluation.

\subsubsection{Text Report Characteristics}
\label{subsubsec:text-characteristics}
The associated clinical reports exhibit substantial variability in descriptive detail and writing style, with an average length of 198 characters (in Chinese). This diversity reflects the nuanced and personalized nature of real-world clinical reporting.

\subsubsection{Dataset Splits}
\label{subsubsec:dataset-splits}
We constructed the test set using stratified sampling based on the primary examination category to ensure balanced evaluation across all major clinical domains. The detailed composition of the test set is summarized in Table~\ref{tab:test-set-composition}. The remaining data was randomly divided into training and validation sets.

\begin{table}[htbp]
\centering
\caption{Composition of the test set by examination category.}
\vspace{6pt}
\label{tab:test-set-composition}
\resizebox{0.4\linewidth}{!}{
\begin{tabular}{lc}
\toprule
\textbf{Examination Type} & \textbf{Number of Cases} \\
\midrule
Upper Abdomen & 331 \\
Thyroid & 200 \\
Breast & 182 \\
Gynecology & 200 \\
Male Urinary & 62 \\
\bottomrule
\end{tabular}
}
\end{table}

\section{Model Training}
We adopt Qwen3-VL-4B-Instruct as the base LVLM and adapt it to ultrasound examination understanding through supervised fine-tuning. Given a multi-image ultrasound examination, the model takes ultrasound images together with a textual instruction as input and is trained to autoregressively generate the corresponding clinical report or target response.

To improve training efficiency while preserving the general multimodal capabilities of the pretrained model, we employ low-rank adaptation (LoRA) for parameter-efficient fine-tuning. Specifically, LoRA adapters are inserted into the self-attention projection layers of the decoder, including \texttt{q\_proj}, \texttt{k\_proj}, \texttt{v\_proj}, and \texttt{o\_proj}, while all remaining pretrained parameters are frozen. The LoRA rank is set to $r=16$, with a scaling factor $\alpha=32$ and a dropout rate of 0.1. The model is optimized using the standard autoregressive cross-entropy loss over the target token sequence.

Training is conducted for 182,936 iterations on NVIDIA GPUs using bfloat16 mixed-precision computation. We optimize the model using AdamW with a learning rate of $1 \times 10^{-4}$, a weight decay of 0.01, and a cosine learning-rate schedule with a warm-up ratio of 0.03. The batch size is set to 8 without additional gradient accumulation. To improve training stability, gradient clipping with a maximum norm of 1.0 is applied. The maximum sequence length is set to 4,096 tokens. For visual inputs, we constrain image resolution by limiting the minimum and maximum numbers of pixels to 784 and 112,896, respectively.

\begin{sidewaystable}
\caption{Performance comparison of general-purpose (\ding{171}), medical-domain (\ding{170}), and ultrasound-specific (\ding{168}) LVLMs across different examination types. BL, RG-L, MTR, BS, and F1 denote BLEU, ROUGE-L, METEOR, BERTScore, and F1 score, respectively. \textbf{Note.} Representative LVLMs, including LLaVA-1.6-Vicuna, Janus-Pro, LLaVA-Med v1.5, HealthGPT, and Med-Flamingo, were excluded due to insufficient long-context support for multi-image ultrasound examinations or unstable outputs when processing a large number of images.}
\vspace{6pt}
\label{tab:comparision_of_sota}
\renewcommand{\arraystretch}{1.0}
\centering
\scriptsize
\begin{minipage}{1.0\textheight}
\centering
\resizebox{\textheight}{!}{%
\setlength{\tabcolsep}{3pt}
\begin{tabular}{c|l|>{\hspace{0.03cm}}cccccccc<{\hspace{0.05cm}}|>{\hspace{0.03cm}}cccccccc<{\hspace{0.05cm}}|>{\hspace{0.03cm}}cccccccc<{\hspace{0.05cm}}}
\toprule
 & & \multicolumn{8}{c|}{\textbf{Thyroid}} & \multicolumn{8}{c|}{\textbf{Breast}} & \multicolumn{8}{c}{\textbf{Upper Abdomen}} \\
 & & \textbf{BL-1} & \textbf{BL-2} & \textbf{BL-3} & \textbf{BL-4} & \textbf{RG-L} & \textbf{MTR} & \textbf{BS} & \textbf{F1} & \textbf{BL-1} & \textbf{BL-2} & \textbf{BL-3} & \textbf{BL-4} & \textbf{RG-L} & \textbf{MTR} & \textbf{BS} & \textbf{F1} & \textbf{BL-1} & \textbf{BL-2} & \textbf{BL-3} & \textbf{BL-4} & \textbf{RG-L} & \textbf{MTR} & \textbf{BS} & \textbf{F1} \\
\midrule
\multicolumn{1}{c|}{\multirow{14}{*}{\ding{171}}} & InternVL3.5-1B-Instruct & 0.257 & 0.160 & 0.107 & 0.075 & 0.252 & 0.219 & 0.724 & 0.287 & 0.179 & 0.122 & 0.083 & 0.053 & 0.294 & 0.249 & 0.716 & 0.040 & 0.109 & 0.060 & 0.037 & 0.020 & 0.278 & 0.171 & 0.695 & 0.064 \\
& InternVL3.5-2B-Instruct & 0.227 & 0.140 & 0.092 & 0.061 & 0.250 & 0.221 & 0.717 & 0.306 & 0.295 & 0.180 & 0.120 & 0.075 & 0.272 & 0.242 & 0.719 & 0.152 & 0.309 & 0.206 & 0.142 & 0.097 & 0.300 & 0.235 & 0.745 & 0.065 \\
& InternVL3.5-4B-Instruct & 0.210 & 0.134 & 0.088 & 0.060 & 0.260 & 0.219 & 0.724 & 0.298 & 0.262 & 0.165 & 0.110 & 0.069 & 0.271 & 0.251 & 0.715 & 0.140 & 0.326 & 0.211 & 0.145 & 0.100 & 0.292 & 0.269 & 0.742 & 0.132 \\
& InternVL3.5-8B-Instruct & 0.323 & 0.207 & 0.136 & 0.089 & 0.261 & 0.240 & 0.744 & 0.334 & 0.345 & 0.234 & 0.164 & 0.107 & 0.319 & 0.279 & 0.744 & 0.127 & 0.377 & 0.255 & 0.182 & 0.131 & 0.320 & 0.286 & 0.763 & 0.054 \\
& InternVL3.5-14B-Instruct & 0.295 & 0.187 & 0.126 & 0.086 & 0.245 & 0.215 & 0.719 & 0.228 & 0.316 & 0.201 & 0.136 & 0.087 & 0.295 & 0.279 & 0.734 & 0.372 & 0.375 & 0.235 & 0.155 & 0.101 & 0.295 & 0.269 & 0.747 & 0.112 \\
& InternVL3.5-30B-A3B-Instruct & 0.331 & 0.210 & 0.139 & 0.095 & 0.256 & 0.237 & 0.728 & 0.285 & 0.352 & 0.228 & 0.153 & 0.096 & 0.340 & 0.274 & 0.747 & 0.380 & 0.351 & 0.243 & 0.174 & 0.125 & 0.343 & 0.291 & 0.764 & 0.079 \\
& InternVL3.5-38B-Instruct & 0.311 & 0.189 & 0.125 & 0.084 & 0.239 & 0.213 & 0.709 & 0.278 & 0.317 & 0.213 & 0.148 & 0.097 & 0.322 & 0.286 & 0.735 & 0.337 & 0.329 & 0.206 & 0.139 & 0.092 & 0.273 & 0.252 & 0.724 & 0.074 \\
& Qwen3.5-0.8B & 0.247 & 0.134 & 0.085 & 0.054 & 0.199 & 0.166 & 0.695 & 0.202 & 0.188 & 0.108 & 0.068 & 0.040 & 0.223 & 0.194 & 0.695 & 0.124 & 0.162 & 0.084 & 0.051 & 0.029 & 0.183 & 0.168 & 0.683 & 0.327 \\
& Qwen3.5-2B & 0.238 & 0.134 & 0.085 & 0.055 & 0.205 & 0.205 & 0.701 & 0.219 & 0.203 & 0.120 & 0.077 & 0.047 & 0.224 & 0.236 & 0.698 & 0.084 & 0.205 & 0.109 & 0.065 & 0.038 & 0.214 & 0.229 & 0.702 & 0.051 \\
& Qwen3.5-4B & 0.229 & 0.125 & 0.075 & 0.044 & 0.203 & 0.220 & 0.706 & 0.178 & 0.181 & 0.110 & 0.071 & 0.043 & 0.248 & 0.265 & 0.710 & 0.467 & 0.210 & 0.115 & 0.066 & 0.037 & 0.231 & 0.257 & 0.714 & 0.228 \\
& Qwen3.5-9B & 0.307 & 0.181 & 0.116 & 0.075 & 0.234 & 0.232 & 0.720 & 0.323 & 0.247 & 0.154 & 0.101 & 0.065 & 0.280 & 0.293 & 0.732 & 0.232 & 0.251 & 0.145 & 0.090 & 0.055 & 0.280 & 0.290 & 0.733 & 0.182 \\
& Qwen3.5-27B & 0.362 & 0.226 & 0.146 & 0.094 & 0.268 & 0.254 & 0.737 & 0.335 & 0.325 & 0.220 & 0.153 & 0.105 & 0.352 & 0.353 & 0.762 & 0.283 & 0.386 & 0.257 & 0.176 & 0.119 & 0.372 & 0.363 & 0.777 & 0.293 \\
& Qwen3.5-35B-A3B & 0.284 & 0.172 & 0.106 & 0.066 & 0.243 & 0.260 & 0.726 & 0.338 & 0.237 & 0.149 & 0.098 & 0.061 & 0.288 & 0.312 & 0.731 & 0.332 & 0.295 & 0.184 & 0.117 & 0.074 & 0.303 & 0.321 & 0.751 & 0.183 \\
& Kimi-VL-A3B-Instruct-16B & 0.057 & 0.035 & 0.023 & 0.016 & 0.264 & 0.189 & 0.716 & 0.235 & 0.133 & 0.094 & 0.066 & 0.044 & 0.322 & 0.261 & 0.736 & 0.273 & 0.401 & 0.282 & 0.201 & 0.146 & 0.363 & 0.303 & 0.776 & 0.163 \\
\midrule
\multicolumn{1}{c|}{\multirow{5}{*}{\ding{170}}} & HuatuoGPT-Vision-7B & 0.084 & 0.054 & 0.039 & 0.027 & 0.237 & 0.142 & 0.703 & 0.457 & 0.239 & 0.153 & 0.106 & 0.067 & 0.260 & 0.221 & 0.717 & 0.137 & 0.147 & 0.098 & 0.070 & 0.049 & 0.275 & 0.205 & 0.737 & 0.109 \\
& HuatuoGPT-Vision-34B & 0.253 & 0.164 & 0.113 & 0.081 & 0.266 & 0.223 & 0.736 & 0.317 & 0.323 & 0.206 & 0.140 & 0.087 & 0.297 & 0.263 & 0.728 & 0.148 & 0.302 & 0.201 & 0.142 & 0.099 & 0.300 & 0.256 & 0.754 & 0.075 \\
& Lingshu-7B & 0.252 & 0.155 & 0.105 & 0.070 & 0.239 & 0.184 & 0.708 & 0.339 & 0.297 & 0.193 & 0.136 & 0.090 & 0.286 & 0.249 & 0.728 & 0.281 & 0.377 & 0.243 & 0.163 & 0.107 & 0.271 & 0.275 & 0.748 & 0.067 \\
& Lingshu-32B & 0.294 & 0.183 & 0.119 & 0.078 & 0.268 & 0.229 & 0.733 & 0.223 & 0.318 & 0.204 & 0.137 & 0.086 & 0.316 & 0.271 & 0.744 & 0.300 & 0.340 & 0.217 & 0.141 & 0.094 & 0.290 & 0.277 & 0.740 & 0.178 \\
& medgemma-4b-it & 0.131 & 0.059 & 0.032 & 0.014 & 0.135 & 0.143 & 0.632 & 0.149 & 0.107 & 0.055 & 0.032 & 0.016 & 0.151 & 0.148 & 0.627 & 0.227 & 0.113 & 0.055 & 0.031 & 0.015 & 0.167 & 0.173 & 0.645 & 0.068 \\
\midrule
\multicolumn{1}{c|}{\multirow{2}{*}{\ding{168}}} & EchoVLM & 0.223 & 0.129 & 0.078 & 0.043 & 0.293 & 0.218 & 0.731 & 0.286 & 0.112 & 0.072 & 0.050 & 0.034 & 0.366 & 0.216 & 0.745 & 0.427 & 0.183 & 0.116 & 0.077 & 0.054 & 0.358 & 0.227 & 0.745 & 0.061 \\
& LUMI & 0.796 & 0.738 & 0.682 & 0.630 & 0.757 & 0.780 & 0.935 & 0.667 & 0.724 & 0.653 & 0.592 & 0.538 & 0.758 & 0.721 & 0.916 & 0.727 & 0.715 & 0.665 & 0.623 & 0.587 & 0.784 & 0.713 & 0.907 & 0.714 \\
\bottomrule
\end{tabular}
}
\end{minipage}
\\
\begin{minipage}{1.0\textheight}
\centering
\resizebox{\linewidth}{!}{%
\setlength{\tabcolsep}{3pt}
\begin{tabular}{c|l|>{\hspace{0.03cm}}cccccccc<{\hspace{0.05cm}}|>{\hspace{0.03cm}}cccccccc<{\hspace{0.05cm}}|>{\hspace{0.03cm}}cccccccc<{\hspace{0.05cm}}}
\toprule
 & & \multicolumn{8}{c|}{\textbf{Gynecology}} & \multicolumn{8}{c|}{\textbf{Male Urinary}} & \multicolumn{8}{c}{\textbf{AVG.}} \\
 & & \textbf{BL-1} & \textbf{BL-2} & \textbf{BL-3} & \textbf{BL-4} & \textbf{RG-L} & \textbf{MTR} & \textbf{BS} & \textbf{F1} & \textbf{BL-1} & \textbf{BL-2} & \textbf{BL-3} & \textbf{BL-4} & \textbf{RG-L} & \textbf{MTR} & \textbf{BS} & \textbf{F1} & \textbf{BL-1} & \textbf{BL-2} & \textbf{BL-3} & \textbf{BL-4} & \textbf{RG-L} & \textbf{MTR} & \textbf{BS} & \textbf{F1} \\
\midrule
\multicolumn{1}{c|}{\multirow{14}{*}{\ding{171}}} & InternVL3.5-1B-Instruct & 0.063 & 0.029 & 0.017 & 0.009 & 0.270 & 0.124 & 0.667 & 0.184 & 0.087 & 0.039 & 0.020 & 0.008 & 0.222 & 0.109 & 0.654 & 0.040 & 0.139 & 0.082 & 0.053 & 0.033 & 0.263 & 0.174 & 0.691 & 0.123 \\
& InternVL3.5-2B-Instruct & 0.137 & 0.085 & 0.058 & 0.039 & 0.273 & 0.211 & 0.712 & 0.173 & 0.212 & 0.124 & 0.086 & 0.062 & 0.224 & 0.152 & 0.693 & 0.023 & 0.236 & 0.147 & 0.100 & 0.067 & 0.264 & 0.212 & 0.717 & 0.144 \\
& InternVL3.5-4B-Instruct & 0.265 & 0.155 & 0.100 & 0.060 & 0.267 & 0.229 & 0.720 & 0.230 & 0.229 & 0.139 & 0.097 & 0.069 & 0.231 & 0.169 & 0.696 & 0.047 & 0.258 & 0.161 & 0.108 & 0.072 & 0.264 & 0.227 & 0.719 & 0.169 \\
& InternVL3.5-8B-Instruct & 0.292 & 0.191 & 0.129 & 0.087 & 0.312 & 0.265 & 0.740 & 0.308 & 0.157 & 0.096 & 0.067 & 0.047 & 0.227 & 0.163 & 0.705 & 0.039 & 0.299 & 0.197 & 0.136 & 0.092 & 0.288 & 0.247 & 0.739 & 0.172 \\
& InternVL3.5-14B-Instruct & 0.318 & 0.196 & 0.126 & 0.082 & 0.302 & 0.269 & 0.741 & 0.187 & 0.231 & 0.140 & 0.094 & 0.064 & 0.265 & 0.184 & 0.708 & 0.011 & 0.307 & 0.192 & 0.127 & 0.084 & 0.280 & 0.243 & 0.730 & 0.182 \\
& InternVL3.5-30B-A3B-Instruct & 0.331 & 0.205 & 0.132 & 0.085 & 0.327 & 0.266 & 0.760 & 0.232 & 0.209 & 0.131 & 0.088 & 0.058 & 0.285 & 0.186 & 0.725 & 0.058 & 0.315 & 0.203 & 0.137 & 0.092 & 0.310 & 0.251 & 0.745 & 0.207 \\
& InternVL3.5-38B-Instruct & 0.370 & 0.239 & 0.157 & 0.104 & 0.350 & 0.296 & 0.766 & 0.254 & 0.237 & 0.148 & 0.098 & 0.061 & 0.304 & 0.199 & 0.722 & 0.011 & 0.313 & 0.199 & 0.133 & 0.088 & 0.298 & 0.249 & 0.731 & 0.191 \\
& Qwen3.5-0.8B & 0.167 & 0.093 & 0.058 & 0.032 & 0.210 & 0.183 & 0.681 & 0.359 & 0.213 & 0.106 & 0.063 & 0.034 & 0.186 & 0.132 & 0.669 & 0.012 & 0.195 & 0.105 & 0.065 & 0.038 & 0.200 & 0.169 & 0.684 & 0.205 \\
& Qwen3.5-2B & 0.165 & 0.090 & 0.055 & 0.031 & 0.213 & 0.211 & 0.680 & 0.121 & 0.248 & 0.130 & 0.077 & 0.043 & 0.217 & 0.184 & 0.686 & 0.022 & 0.212 & 0.117 & 0.072 & 0.043 & 0.215 & 0.213 & 0.693 & 0.099 \\
& Qwen3.5-4B & 0.149 & 0.081 & 0.048 & 0.027 & 0.249 & 0.243 & 0.706 & 0.384 & 0.304 & 0.162 & 0.093 & 0.051 & 0.238 & 0.198 & 0.701 & 0.044 & 0.214 & 0.119 & 0.070 & 0.040 & 0.234 & 0.237 & 0.707 & 0.260 \\
& Qwen3.5-9B & 0.224 & 0.127 & 0.076 & 0.046 & 0.273 & 0.270 & 0.720 & 0.201 & 0.288 & 0.156 & 0.096 & 0.062 & 0.261 & 0.217 & 0.710 & 0.172 & 0.263 & 0.153 & 0.096 & 0.061 & 0.265 & 0.261 & 0.723 & 0.222 \\
& Qwen3.5-27B & 0.274 & 0.163 & 0.099 & 0.061 & 0.331 & 0.306 & 0.746 & 0.224 & 0.378 & 0.230 & 0.152 & 0.102 & 0.355 & 0.275 & 0.747 & 0.402 & 0.345 & 0.220 & 0.145 & 0.096 & 0.335 & 0.310 & 0.754 & 0.307 \\
& Qwen3.5-35B-A3B & 0.241 & 0.143 & 0.088 & 0.054 & 0.287 & 0.284 & 0.730 & 0.207 & 0.318 & 0.182 & 0.111 & 0.069 & 0.312 & 0.267 & 0.733 & 0.370 & 0.275 & 0.166 & 0.104 & 0.065 & 0.287 & 0.289 & 0.734 & 0.286 \\
& Kimi-VL-A3B-Instruct-16B & 0.313 & 0.211 & 0.148 & 0.104 & 0.333 & 0.265 & 0.749 & 0.322 & 0.213 & 0.132 & 0.088 & 0.056 & 0.264 & 0.171 & 0.711 & 0.069 & 0.223 & 0.151 & 0.105 & 0.073 & 0.309 & 0.238 & 0.738 & 0.212 \\
\midrule
\multicolumn{1}{c|}{\multirow{5}{*}{\ding{170}}} & HuatuoGPT-Vision-7B & 0.258 & 0.164 & 0.114 & 0.080 & 0.301 & 0.243 & 0.728 & 0.200 & 0.120 & 0.072 & 0.045 & 0.027 & 0.240 & 0.143 & 0.699 & 0.127 & 0.170 & 0.108 & 0.075 & 0.050 & 0.263 & 0.191 & 0.717 & 0.206 \\
& HuatuoGPT-Vision-34B & 0.273 & 0.167 & 0.110 & 0.070 & 0.277 & 0.238 & 0.729 & 0.196 & 0.177 & 0.109 & 0.072 & 0.046 & 0.283 & 0.177 & 0.711 & 0.109 & 0.266 & 0.169 & 0.115 & 0.076 & 0.285 & 0.232 & 0.732 & 0.169 \\
& Lingshu-7B & 0.290 & 0.184 & 0.124 & 0.084 & 0.288 & 0.265 & 0.742 & 0.201 & 0.162 & 0.092 & 0.058 & 0.036 & 0.219 & 0.143 & 0.691 & 0.120 & 0.276 & 0.173 & 0.117 & 0.078 & 0.260 & 0.223 & 0.723 & 0.202 \\
& Lingshu-32B & 0.341 & 0.219 & 0.149 & 0.104 & 0.309 & 0.272 & 0.753 & 0.146 & 0.209 & 0.116 & 0.074 & 0.046 & 0.211 & 0.145 & 0.679 & 0.122 & 0.300 & 0.188 & 0.124 & 0.081 & 0.279 & 0.239 & 0.730 & 0.194 \\
& medgemma-4b-it & 0.129 & 0.059 & 0.033 & 0.014 & 0.169 & 0.159 & 0.636 & 0.181 & 0.154 & 0.066 & 0.035 & 0.015 & 0.143 & 0.127 & 0.622 & 0.000 & 0.127 & 0.059 & 0.033 & 0.015 & 0.153 & 0.150 & 0.632 & 0.125 \\
\midrule
\multicolumn{1}{c|}{\multirow{2}{*}{\ding{168}}} & EchoVLM & 0.143 & 0.081 & 0.049 & 0.030 & 0.333 & 0.206 & 0.729 & 0.335 & 0.197 & 0.114 & 0.072 & 0.043 & 0.268 & 0.163 & 0.704 & 0.058 & 0.172 & 0.102 & 0.065 & 0.041 & 0.323 & 0.206 & 0.731 & 0.233 \\
& LUMI & 0.643 & 0.579 & 0.527 & 0.482 & 0.734 & 0.637 & 0.894 & 0.566 & 0.875 & 0.858 & 0.840 & 0.821 & 0.912 & 0.874 & 0.948 & 0.850 & 0.751 & 0.699 & 0.653 & 0.612 & 0.789 & 0.745 & 0.920 & 0.705 \\
\bottomrule
\end{tabular}
}
\end{minipage}
\end{sidewaystable}

\section{Experiments}
\subsection{Baselines}
We compare LUMI against a diverse set of state-of-the-art LVLMs, including both general-purpose and medical-domain models. The baselines include the InternVL3.5 series (1B--38B) \cite{DBLP:journals/corr/abs-2508-18265}, the Qwen3.5 series (0.8B--35B) \cite{DBLP:journals/corr/abs-2511-21631}, Kimi-VL \cite{DBLP:journals/corr/abs-2504-07491}, HuatuoGPT-Vision (7B and 34B) \cite{DBLP:journals/corr/abs-2406-19280}, Lingshu (7B and 32B) \cite{DBLP:journals/corr/abs-2506-07044}, MedGemma \cite{DBLP:journals/corr/abs-2604-05081}, and EchoVLM \cite{DBLP:journals/corr/abs-2509-14977}, a publicly available ultrasound-specific vision-language model. We additionally considered several other representative LVLMs, including the LLaVA-1.6-Vicuna series (7B and 13B) \cite{liu2024improved}, the Janus-Pro series (1B and 7B) \cite{chen2025janus}, LLaVA-Med v1.5 \cite{DBLP:conf/nips/LiWZULYNPG23}, HealthGPT \cite{lin2025healthgpt}, and Med-Flamingo \cite{DBLP:conf/ml4h/MoorHWYDLZRR23}. These models were excluded from the final comparison because they either do not support the long input sequences required by multi-image ultrasound examinations or produce unstable outputs (e.g., garbled text) when processing a large number of images.

\subsection{Main Results}
We conduct a comprehensive evaluation across five examination categories: Thyroid, Breast, Upper Abdomen, Gynecology, and Male Urinary. Following prior report generation studies, we adopt BLEU-1/2/3/4, ROUGE-L, METEOR, and BERTScore as evaluation metrics.

To further assess clinical correctness, we introduce the F1 score. Specifically, both the generated report and the corresponding ground-truth report are provided to an LLM-based evaluator, which evaluates the consistency of lesion identification and diagnostic conclusions between the two reports. Precision, recall, and F1 scores are then computed on a per-lesion basis, providing a more direct measure of clinical reliability than conventional text-similarity metrics.

The quantitative results are summarized in Table~\ref{tab:comparision_of_sota}. LUMI achieves state-of-the-art performance across all five examination categories and consistently outperforms the competing models on both natural language generation and clinical metrics. Notably, LUMI ranks first across all evaluation metrics in the average (AVG.) column. The improvements are particularly pronounced on higher-order natural language generation metrics and the F1 score, indicating that LUMI not only produces linguistically accurate reports but also captures clinically relevant findings with greater fidelity.

\subsection{Qualitative Analysis}
We provide representative case studies in the Appendix to compare the report generation capabilities of LUMI with those of leading baseline models. The examples illustrate that LUMI generates reports that are more complete, clinically coherent, and better aligned with the underlying ultrasound findings.

\subsection{Discussion}
\subsubsection{Comparison with Medical-Domain Models}
LUMI consistently outperforms medical-domain LVLMs, including HuatuoGPT-Vision, Lingshu, and EchoVLM. These results suggest that large-scale ultrasound-specific supervision plays a crucial role in developing clinically effective ultrasound understanding models, beyond the benefits provided by general medical pretraining alone.

\subsubsection{Clinical Fidelity}
While several baseline models are capable of generating fluent and well-structured reports, they frequently omit clinically important findings or introduce unsupported diagnostic statements. In contrast, LUMI demonstrates stronger factual consistency with the reference reports and better adherence to clinical reporting conventions, leading to improved diagnostic reliability.

\subsubsection{Model Scaling}
\begin{figure}[t]
\centering
\includegraphics[width=0.86\textwidth]{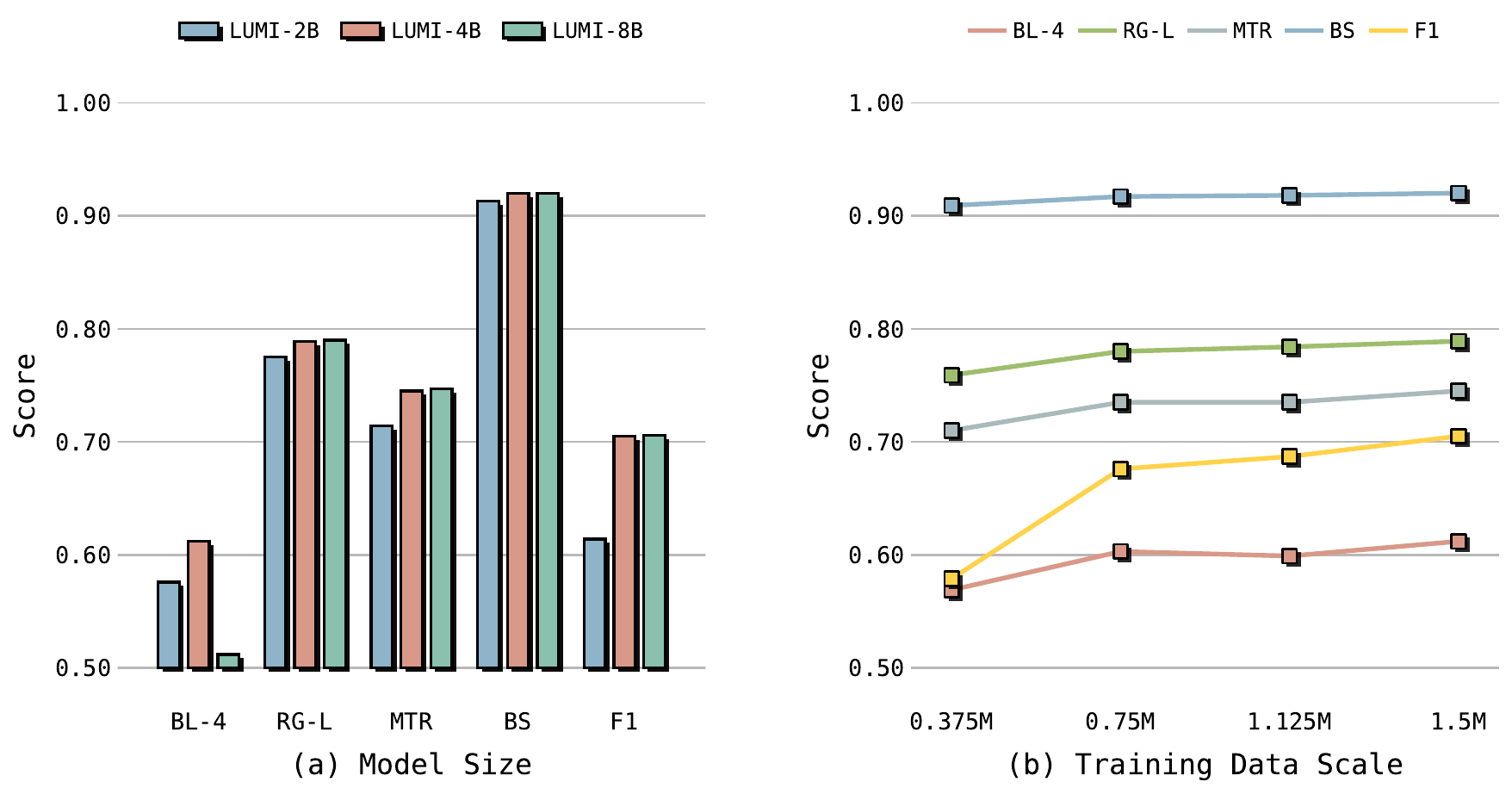}
\caption{Performance under different model and data scales. (a) Model size. (b) Training data scale. BL, RG-L, MTR, BS, and F1 denote BLEU, ROUGE-L, METEOR, BERTScore, and F1 score, respectively.}
\label{fig:scaling-comparison}
\end{figure}

We further investigate the impact of model scale. Under the same experimental setting and training recipe, we train 2B and 8B variants and compare them with the 4B model. As shown in Figure~\ref{fig:scaling-comparison}, the 2B model consistently underperforms the 4B model across all evaluation metrics, indicating that increasing model capacity benefits ultrasound understanding. The improvement is particularly pronounced in terms of the F1 score: the 2B model achieves an F1 score of 0.614, whereas the 4B model reaches 0.705, corresponding to an absolute gain of 0.091. However, scaling the model further from 4B to 8B yields no substantial performance improvement across the evaluation metrics. This suggests that performance gains saturate beyond the 4B scale under our current training setup and dataset.

\subsubsection{Data Scaling}
We also analyze the effect of data scale by training models using 25\%, 50\%, 75\%, and 100\% of the training data. As shown in Figure~\ref{fig:scaling-comparison}, performance consistently improves as the amount of training data increases, highlighting the importance of large-scale, domain-specific data for ultrasound LVLMs. Interestingly, even with a relatively small training set, the model already achieves competitive scores on natural language generation metrics such as BERTScore. In contrast, the clinical F1 score remains substantially lower, suggesting that the generated reports may be linguistically and structurally plausible while still lacking sufficient diagnostic accuracy. Notably, at the largest training scale evaluated (approximately 1.5 million samples), the performance curves for most evaluation metrics---particularly the clinical F1 score---show no clear sign of saturation. This finding suggests that further scaling the training data could yield additional performance gains.

\subsubsection{Limitation}
Since LUMI is trained on paired data consisting of all images from a complete ultrasound examination and the corresponding report, it may hallucinate findings when presented with severely incomplete image sets at inference time.

\section{Conclusion}
In this work, we present LUMI, an LVLM specialized for real-world ultrasound understanding. To support this effort, we construct a large-scale ultrasound dataset consisting of nearly 1.5 million examinations and 17.7 million images spanning multiple anatomical regions, with examination-level image-text alignment that reflects real clinical workflows. Experimental results demonstrate that LUMI achieves state-of-the-art performance across a wide range of natural language generation and clinical metrics, outperforming both general-purpose and medical-domain LVLMs. Furthermore, our model and data scaling analyses provide insights into the role of scale in advancing LVLMs for ultrasound and other medical imaging applications.

\bibliographystyle{unsrt}
\bibliography{references}

\includepdf[
    pages=1,
    width=0.95\textwidth,
    pagecommand={
        \section*{Appendix}
        \thispagestyle{plain}
    }
]{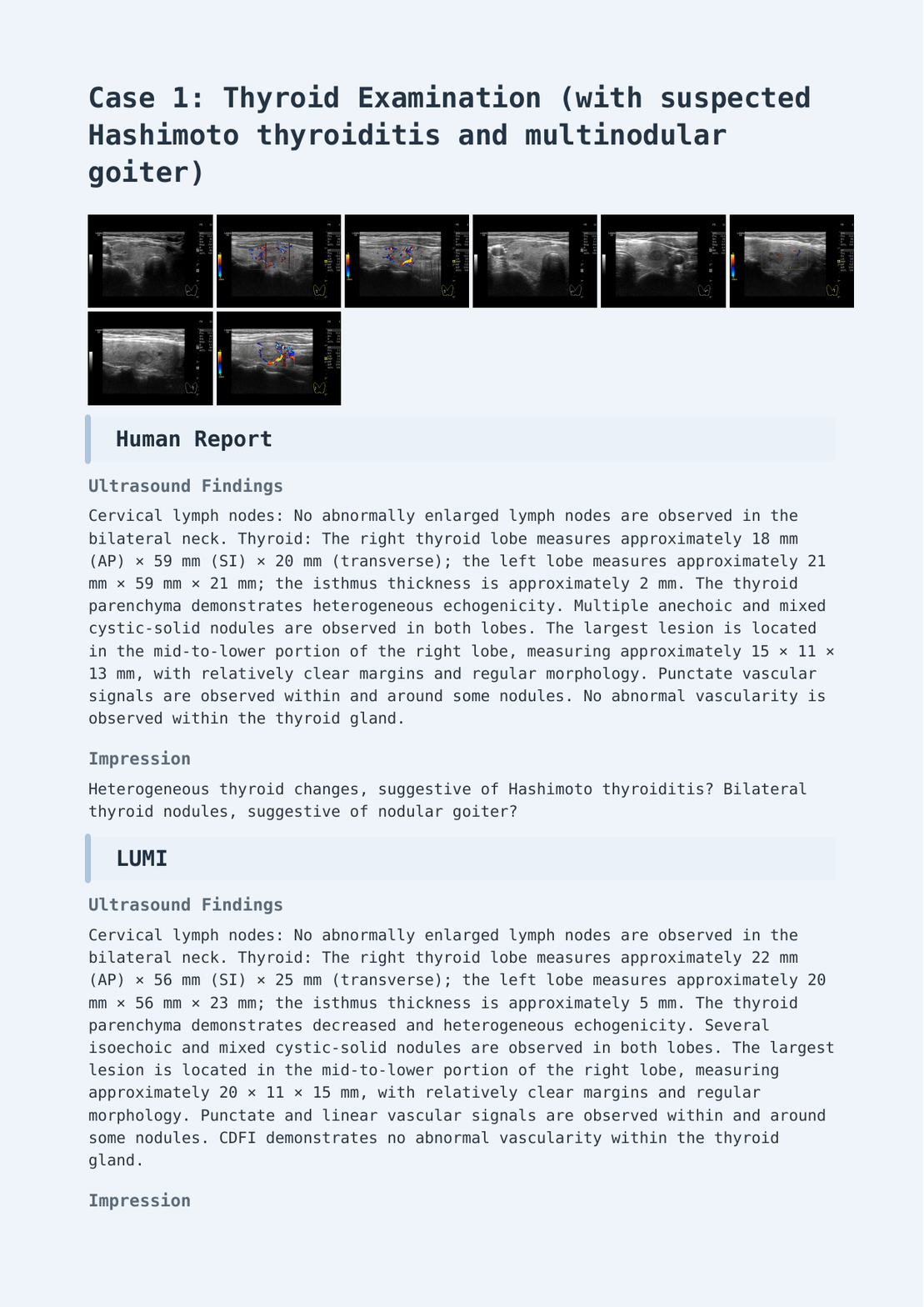}

\includepdf[
    pages=2-,
    width=0.95\textwidth
]{figures/qualitative_cases_20260630.pdf}

\end{document}